\documentclass[10pt,twocolumn,letterpaper]{article}
\pdfoutput=1

\usepackage{cvpr}
\usepackage{times}
\usepackage{epsfig}
\usepackage{graphicx}
\usepackage{amsmath}
\usepackage{amssymb}

\usepackage{algorithm}
\usepackage{algorithmic}
\usepackage{mathtools}
\usepackage{array}
\usepackage{multirow}
\usepackage{booktabs}
\usepackage{makecell}
\usepackage{colortbl}
\usepackage{textcomp} 
\usepackage[pagebackref=true,breaklinks=true,letterpaper=true,colorlinks,bookmarks=false]{hyperref}

\cvprfinalcopy 

\def\cvprPaperID{1558} 

\ifcvprfinal\fi
\begin{document}

\title{Unifying Heterogeneous Classifiers with Distillation}

\author{Jayakorn Vongkulbhisal$^1$, Phongtharin Vinayavekhin$^1$, Marco Visentini-Scarzanella$^{2}$\\
$^1$IBM Research, Tokyo, Japan\\
$^2$Amazon, Tokyo, Japan\\
{\tt\small jayakornv@ibm.com, pvmilk@jp.ibm.com, marcovs@amazon.com}
}

\maketitle

\newcommand{\nclsfs}{N}
\newcommand{\nclasses}{L}
\newcommand{\minimize}{\text{minimise}}
\renewcommand{\algorithmicrequire}{\textbf{Input:}}
\renewcommand{\algorithmicensure}{\textbf{Output:}}
\newcommand\myeq{\mkern3.5mu{=}\mkern5mu}

\begin{abstract}
In this paper, we study the problem of unifying knowledge from a set of classifiers with different architectures and target classes into a single classifier, given only a generic set of unlabelled data. We call this problem Unifying Heterogeneous Classifiers (UHC). This problem is motivated by scenarios where data is collected from multiple sources, but the sources cannot share their data, \eg, due to privacy concerns, and only privately trained models can be shared. In addition, each source may not be able to gather data to train all classes due to data availability at each source, and may not be able to train the same classification model due to different computational resources. To tackle this problem, we propose a generalisation of knowledge distillation to merge HCs. We derive a probabilistic relation between the outputs of HCs and the probability over all classes. Based on this relation, we propose two classes of methods based on cross-entropy minimisation and matrix factorisation, which allow us to estimate soft labels over all classes from unlabelled samples and use them in lieu of ground truth labels to train a unified classifier. Our extensive experiments on ImageNet, LSUN, and Places365 datasets show that our approaches significantly outperform a naive extension of distillation and can achieve almost the same accuracy as classifiers that are trained in a centralised, supervised manner.
\end{abstract}


\section{Introduction}

\begin{figure}
	\centering
	\includegraphics[width=1\linewidth]{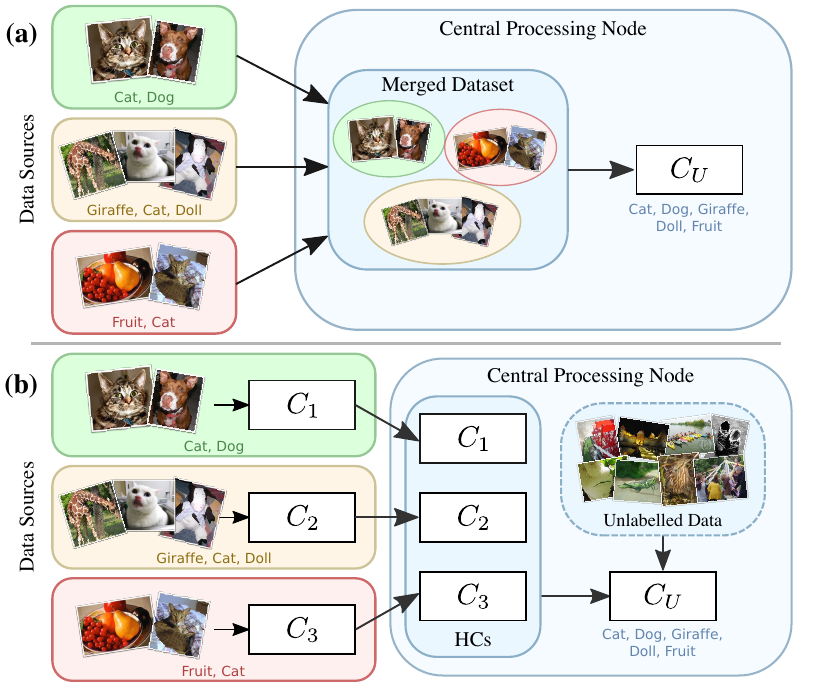}
	\caption{Unifying Heterogeneous Classifiers. (a) Common training approaches require transferring data from sources to a central processing node where a classifier is trained. (b) We propose to train a unified classifier from pre-trained classifiers from each source and an unlabelled set of generic data, thereby preserving privacy. The individual pre-trained classifiers may have different sets of target classes, hence the term \textit{Heterogeneous Classifiers} (HCs).}
	\label{fig:first_fig}	
\end{figure}

The success of machine learning in image classification tasks has been largely enabled by the availability of big datasets, such as ImageNet~\cite{Russakovsky2015} and MS-COCO~\cite{Lin2014}. As the technology becomes more pervasive, data collection is transitioning towards more distributed settings where the data is sourced from multiple entities and then combined to train a classifier in a central node (Fig.~\ref{fig:first_fig}a). However, in many cases, transfer of data between entities is not possible due to privacy concerns (\eg, private photo albums or medical data) or bandwidth restrictions (\eg, very large datasets), hampering the unification of knowledge from different sources. This has led to multiple works that propose to learn classifiers without directly sharing data, \eg, distributed optimisation~\cite{Boyd2011}, consensus-based training~\cite{Forero2010}, and federated learning~\cite{Konecny2016}. However, these approaches generally require models trained by each entity to be the same both in terms of architecture and target classes.

In this paper, we aim to remove these limitations and propose a system for a more general scenario consisting of an ensemble of Heterogeneous Classifiers (HCs), as shown in Fig.~\ref{fig:first_fig}b. We define a set of HCs as a set of classifiers which may have different architectures and, more importantly, may be trained to classify different sets of target classes. To combine the HCs, each entity only needs to forward their trained classifiers and class names to the central processing node, where all the HCs are unified into a single model that can classify all target classes of all input HCs. We refer to this problem as \textit{Unifying Heterogeneous Classifiers} (UHC). UHC has practical applications for the cases when it is not possible to enforce every entity to (\textit{i}) use the same model/architecture; (\textit{ii}) collect sufficient training data for all classes; or (\textit{iii}) send the data to the central node, due to computational, data availability, and confidentiality constraints.

To tackle UHC, we propose a generalisation of knowledge distillation \cite{Bucila2006, Hinton2015}. Knowledge distillation was originally proposed to compress multiple complex \textit{teacher} models into a single simpler \textit{student} one. However, distillation still assumes that the target classes of all teacher and student models are the same, whereas in this work we relax this limitation. To generalise distillation to UHC, we derive a probabilistic relationship connecting the outputs of HCs with that of the unified classifier. Based on this relationship, we propose two classes of methods, one based on cross-entropy minimisation and the other on matrix factorisation with missing entries, to estimate the probability over all classes of a given sample. After obtaining the probability, we can then use it to train the unified classifier. Our approach only requires unlabelled data to unify HCs, thus no labour is necessary to label any data at the central node. In addition, our approach can be applied to any classifiers which can be trained with soft labels, \eg, neural networks, boosting classifiers, random forests, \etc.

We evaluated our proposed approach extensively on ImageNet, LSUN, and Places365 datasets in a variety of settings and against a natural extension of the standard distillation. Through our experiments we show that our approach outperforms standard distillation and can achieve almost the same accuracy as the classifiers that were trained in a centralised, supervised manner.

\section{Related Work}
There exists a long history of research that aims to harness the power of multiple classifiers to boost classification result. The most well-known approaches are arguably ensemble methods~\cite{Kittler1998, Kuncheva2004, Polikar2006} which combine the output of multiple classifiers to make a classification. Many techniques, such as voting and averaging~\cite{Kuncheva2004}, can merge prediction from trained classifiers, while some train the classifiers jointly as part of the technique, \eg, boosting~\cite{Freund1999} and random forests~\cite{Breiman2001}. These techniques have been successfully used in many applications,~\eg, multi-class classification~\cite{Hastie2009}, object detection~\cite{Viola2001,Malisiewicz2011}, tracking~\cite{Avidan2007},~\etc. However, ensemble methods require storing and running all models for prediction, which may lead to scalability issues when complex models,~\eg, deep networks, are used. In addition, ensemble methods assume all base classifiers are trained to classify all classes, which is not suitable for the scenarios addressed by UHC.

To the best of our knowledge, the closest class of methods to UHC is knowledge distillation~\cite{Bucila2006, Hinton2015}. Distillation approaches operate by passing unlabelled data to a set of pretrained {\it teacher} models to obtain soft predictions, which are used to train a {\it student} model. 
Albeit originally conceived for compressing complex models into simpler ones by matching predictions, distillation has been further extended to, for instance, matching intermediate features~\cite{Romero2015}, knowledge transfer between domains~\cite{Gupta2016}, combining knowledge using generative adversarial-based loss~\cite{Xu2017},~\etc. More related to UHC,  Lopes~\etal~\cite{Lopes2017} propose to distill teacher models trained by different entities using their metadata rather than raw inputs. This allows the student model to be trained without any raw data transfer, thus preserving privacy while also not requiring any data collection from the central processing node. Still, no formulation of distillation can cope with the case where each teacher model has different target classes, which we tackle in this paper. We describe how distillation can be generalised to UHC in the next section.

\section{Unifying Heterogeneous Classifiers (UHC)}
\label{sec:method}
\begin{figure}
\centering
\includegraphics[width=1\linewidth]{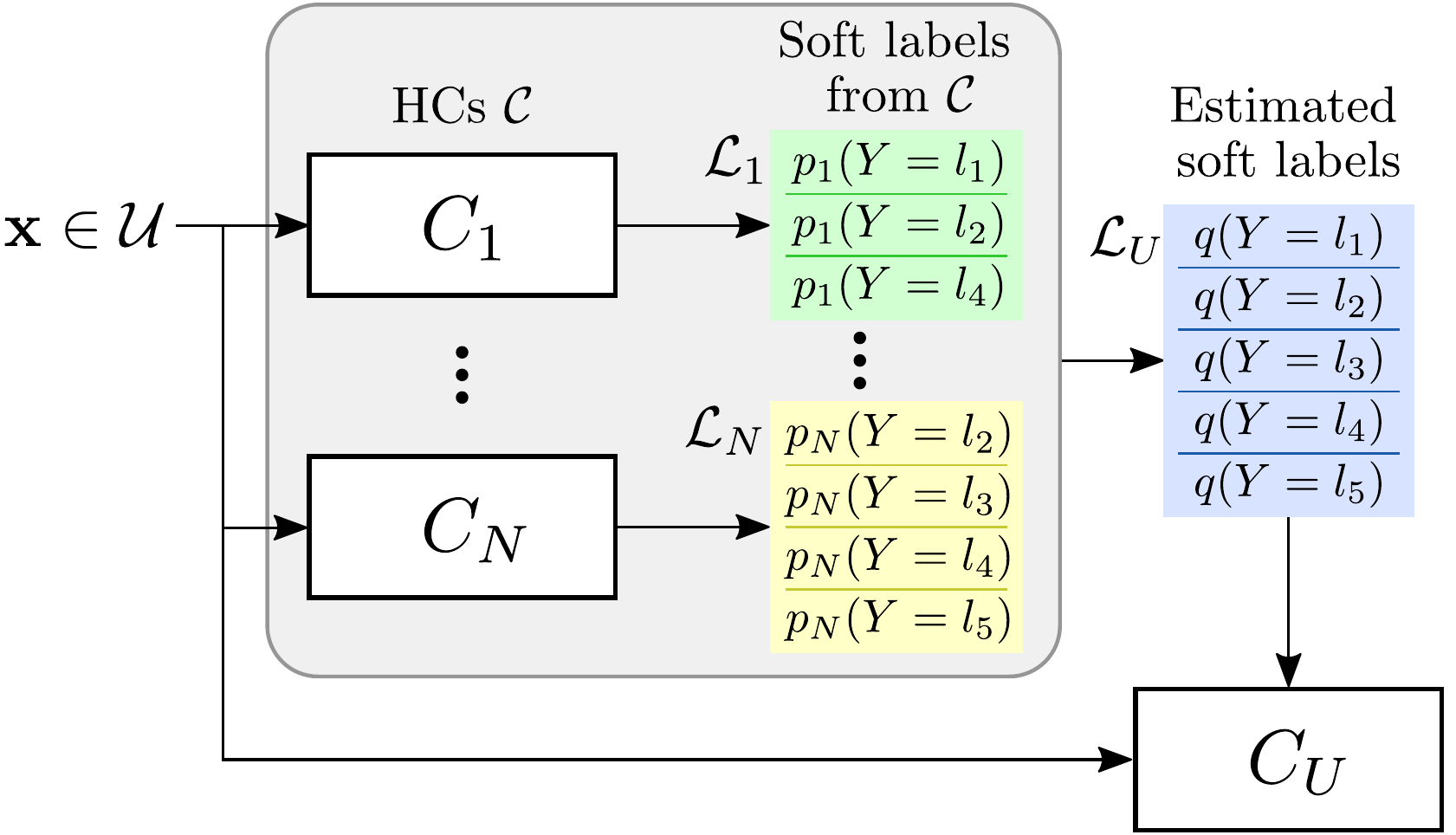}
\caption{UHC problem and approach overview. An input image $\mathbf{x}$ is drawn from an unlabelled set $\mathcal{U}$ and input to a set of pre-trained classifiers $\{C_1,\cdots,C_N\}$, where each $C_i$ returns soft label $p_i$ over classes in $\mathcal{L}_i$. Here, the classes $\mathcal{L}_i$ may be different for each $C_i$. The goal of UHC is to train a classifier $C_U$ that can classify all target classes in  $\mathcal{L}_U$ using the prediction of $C_i$ on $\mathbf{x}\in\mathcal{U}$ instead of labelled data. Our approach to UHC involves using $p_i$ to estimate $q$, the soft label of $\mathbf{x}$ over all classes in $\mathcal{L}_U$, then using $\mathbf{x}$ and $q$ to train $C_U$.}
\label{Fig_overview}	
\end{figure}

We define the Unifying Heterogeneous Classifiers (UHC) problem in this paper as follows (see Fig.~\ref{Fig_overview}). Let $\mathcal{U}$ be an unlabelled set of images (``transfer set") and let $\mathcal{C}=\{C_i\}_{i=1}^{\nclsfs}$ be a set of $N$ Heterogeneous Classifiers (HCs), where each $C_i$ is trained to predict the probability $p_i(Y=l_j)$ of an image belonging to class $\l_j\in\mathcal{L}_i$.
Given $\mathcal{U}$ and $\mathcal{C}$, the goal of this work is to learn a \textit{unified classifier} $C_U$ that estimates the probability $q(Y=l_j)$ of an input image belonging to the class $l_j\in\mathcal{L}_U$ where $\mathcal{L}_U=\bigcup_{i=1}^\nclsfs\mathcal{L}_i=\{l_1,l_2,\dots,l_\nclasses\}$. Note that $C_i$ might be trained to classify different sets of classes, \ie, we may have $\mathcal{L}_i \neq \mathcal{L}_j$ or even $\lvert\mathcal{L}_i\rvert \neq \lvert\mathcal{L}_j\rvert$ for $i\neq j$. 

Our approach to tackle UHC involves three steps: ({\it i})~passing the image $\mathbf{x}\in\mathcal{U}$ to $C_i$ to obtain $p_i, \forall i$, ({\it ii})~estimating $q$ from $\{p_i\}_i$, then ({\it iii})~using the estimated $q$ to train $C_U$ in a supervised manner. We note that it is possible to combine ({\it ii}) and {(\it iii)} into a single step for neural networks (see Sec.~\ref{sec:method:DirectBP}), but this 3-step approach allows it to be applied to other classifiers, \eg, boosting and random forests. To accomplish ({\it ii}), we derive probabilistic relationship between each $p_i$ and $q$, which we leverage to estimate $q$ via the following two proposed methods: cross-entropy minimisation and matrix factorisation. 
In the rest of this section, we first review standard distillation, showing why it cannot be applied to UHC. We then describe our approaches to estimate $q$ from $\{p_i\}_i$. We provide a discussion on the computation cost in the supplementary material.

\subsection{Review of Distillation}
\label{sec:ReviewDistillation}
\textbf{Overview} Distillation \cite{Bucila2006, Hinton2015} is a class of algorithms used for compressing multiple trained models $C_i$ into a single unified model $C_U$ using a set of unlabelled data $\mathcal{U}$\footnote{Labelled data can also be used in a supervised manner.}. Referring to Fig.~\ref{Fig_overview}, standard distillation corresponds to the case where $\mathcal{L}_i=\mathcal{L}_j$, $\forall(i,j)$. 
The unified $C_U$ is trained by minimising the cross-entropy between outputs of $C_i$ and $C_U$ as
\begin{equation}
J(q) = -\sum_i \sum_{l\in\mathcal{L}_U} p_i(Y=l) \log q(Y=l).
\label{eq:CrossEntropy_Distillation}
\end{equation}
Essentially, the outputs of $C_i$ are used as soft labels for the unlabelled $\mathcal{U}$ in training $C_U$. For neural networks, class probabilities are usually computed with softmax function:
\begin{equation}
p(Y=l)=\frac{\exp(z_l/T)}{\sum_{k\in\mathcal{L}_U}\exp(z_k/T)},\label{eq:softmax}
\end{equation}
where $z_l$ is the logit for class $l$, and $T$ denotes an adjustable temperature parameter. In \cite{Hinton2015}, it was shown that minimising \eqref{eq:CrossEntropy_Distillation} when $T$ is high is similar to minimising the $\ell_2$ error between the logits of $p$ and $q$, thereby relating the cross-entropy minimising to logit matching.

\textbf{Issues} The main issue with standard distillation stems from its inability to cope with the more general case of $\mathcal{L}_i \neq \mathcal{L}_j$. Mathematically, Eq.~\eqref{eq:CrossEntropy_Distillation} assumes $C_U$ and $C_i$'s share the same set of classes. This is not true in our case since each $C_i$ is trained to predict classes in $\mathcal{L}_i$, thus $p_i(Y=l)$ is undefined for $l \in \mathcal{L}_{-i}$\footnote{We define $\mathcal{L}_{-i}$ as the set of classes in $\mathcal{L}_U$ but outside $\mathcal{L}_i$.}. A naive solution to this issue would be to simply set $p_i(Y=l)=0$ for $l\in \mathcal{L}_{-i}$. However, this could incur serious errors, \eg, one may set $p_i(Y=\text{cat})$ of a cat image to zero when $C_i$ does not classify cats, which would be an improper supervision. We show that this approach does not provide good results in the experiments. 

It is also worth mentioning that $C_i$ in UHC is different from the \textit{Specialised Classifiers} (SC) in~\cite{Hinton2015}. While SCs are trained to specialise in classifying a subset of classes, they are also trained with data from other classes which are grouped together into a single \textit{dustbin class}. This allows SCs to distinguish dustbin from their specialised classes, enabling student model to be trained with~\eqref{eq:CrossEntropy_Distillation}. Using the previous example, the cat image would be labelled as dustbin class, which is an appropriate supervision for SCs that do not classify cat. However, the presence of a dustbin class imposes a design constraint on the $C_i$'s, as well as requiring the data source entities to collect large amounts of generic data to train it. Conversely, we remove these constraints in our formulation, and $C_i$'s are trained without a dustbin class. Thus, given data from $\mathcal{L}_{-i}$, $C_i$ will only provide $p_i$ only over classes in $\mathcal{L}_i$, making it difficult to unify $\mathcal{C}$ with~\eqref{eq:CrossEntropy_Distillation}.

\subsection{Relating outputs of HCs and unified classifier}

To overcome the limitation of standard distillation, we need to relate the output $p_i$ of each $C_i$ to the probability $q$ over $\mathcal{L}_U$. Since $p_i$ is defined only in the subset $\mathcal{L}_i\subseteq\mathcal{L}_U$, we can consider $p_i(Y=l)$ as the probability $q$ of $Y= l$ given that $Y$ cannot be in $\mathcal{L}_{-i}$. This leads to the following derivation:
\begin{align}
p_i(Y=l) &= q(Y=l | Y\notin \mathcal{L}_{-i}) \\
&= q(Y=l | Y\in \mathcal{L}_i) \\
&= \frac{q(Y=l , Y\in \mathcal{L}_i)}{q(Y\in \mathcal{L}_i)} \\
&= \frac{q(Y=l)}{\sum_{k\in\mathcal{L}_i} q(Y=k)}.
\label{eq:relation_pi_and_q}
\end{align}
We can see that $p_i(Y=l)$ is equivalent to $q(Y=l)$ normalised by the classes in $\mathcal{L}_i$. In the following sections, we describe two classes of methods that utilise this relationship for estimating $q$ from $\{p_i\}_i$. 

\subsection{Method 1: Cross-entropy approach}
\label{sec:method1}
Recall that the goal of~\eqref{eq:CrossEntropy_Distillation} is to match $q$ to $p_i$ by minimising the cross-entropy between them. Based on the relation in~\eqref{eq:relation_pi_and_q}, we generalise~\eqref{eq:CrossEntropy_Distillation} to tackle UHC by matching $\frac{q(Y=l)}{\sum_{k\in\mathcal{L}_i} q(Y=k)}$ to $p_i(Y=k)$, resulting in:
\begin{equation}
J(q) = -\sum_i \sum_{l\in\mathcal{L}_i} p_i(Y=l) \log \hat{q}_i(Y=l),
\label{eq:CrossEntropy_Unify}
\end{equation}
where:
\begin{equation}
\hat{q}_i(Y=l) = \frac{q(Y=l)}{\sum_{k\in\mathcal{L}_i}q(Y=k)}.
\end{equation}
We can see that the difference between \eqref{eq:CrossEntropy_Distillation} and \eqref{eq:CrossEntropy_Unify} lies in the normalisation of $q$. Specifically, the cross-entropy of each $C_i$ (\ie, the second summation) is computed between $p_i(Y=l)$ and $\hat{q}_i(Y=l)$ over the classes in $\mathcal{L}_i$. With this approach, we do not need to arbitrarily define values for $p_i(Y=l)$ whenever  $l\in\mathcal{L}_{-i}$, thus not causing spurious supervision. We now outline optimality properties of~\eqref{eq:CrossEntropy_Unify}.

\vspace{4pt}

\noindent\textbf{Proposition 1 (Sufficient condition for optimality)} Suppose there exists a probability $\bar{p}$ over $\mathcal{L}_U$, where $p_i(Y\mkern0.5mu{=}\mkern2mu l)=\frac{\bar{p}(Y=l)}{\sum_{k\in\mathcal{L}_i}\bar{p}(Y=k)},\forall i$, then $q=\bar{p}$ is a global minimum of \eqref{eq:CrossEntropy_Unify}.

\noindent\textbf{Sketch of proof} Consider $\tilde{J}_i(\tilde{q}_i) = -\sum_{l\in\mathcal{L}_i}p_i(Y=l)\log \tilde{q}_i(Y=l)$ (Note $\tilde{J}_i$ is a function of $\tilde{q}_i$ whereas $J$ is a function of $q$). $\tilde{J}_i(\tilde{q}_i)$ achieves its minimum when $\tilde{q}_i=p_i$, with the a value of $\tilde{J}_i(p_i)$. Thus, the minimum value of $\sum_i\tilde{J}_i(\tilde{q}_i)$ is $\sum_i\tilde{J}_i(p_i)$. This is a lower bound of \eqref{eq:CrossEntropy_Unify},~\ie, $\sum_i\tilde{J}_i(p_i)\leq J(q), \forall q$. However, we can see that by setting $q=\bar{p}$, we achieve equality in the bound,~\ie, $\sum_i\tilde{J}_i(p_i)=J(\bar{p})$, and so $\bar{p}$ is a global minimum of~\eqref{eq:CrossEntropy_Unify}. \hfill $\square$ 
\vspace{4pt}

The above result establishes the form of a global minimum of~\eqref{eq:CrossEntropy_Unify}, and that minimising~\eqref{eq:CrossEntropy_Unify} may obtain the true underlying probability $\bar{p}$ if it exists. However, there are cases where the global solution may not be unique. A simple example is when there are no shared classes between the HCs,~\eg, $\nclsfs=2$ with $\mathcal{L}_1\cap\mathcal{L}_2=\emptyset$. It may be possible to show uniqueness of the global solution in some cases depending on the structure of shared classes between $\mathcal{L}_i$'s, but we leave this as future work.

\textbf{Optimisation} Minimisation of~\eqref{eq:CrossEntropy_Unify} can be transformed into a geometric program (see supplementary material), which can then be converted to a convex problem and efficiently solved~\cite{Boyd2007}. In short, we define $u_l\in\mathbb{R}$ for $l\in\mathcal{L}_U$ and replace $q(Y=l)$ with $\exp(u_l)$. Thus,~\eqref{eq:CrossEntropy_Unify} transforms to
\begin{equation}
\footnotesize
\hat{J}(\{u_l\}_l) = -\sum_i \sum_{l\in\mathcal{L}_i} p_i(Y=l) \left(
u_l-\log
\left(
\sum_{k\in\mathcal{L}_i}\exp(u_k)
\right)
\right),
\label{eq:CrossEntropy_Unify_Optimise}
\end{equation}
which is convex in $\{u_l\}_l$ since it is a sum of scaled and log-sum-exps of $\{u_l\}_l$~\cite{Boyd2004}. We minimise it using gradient descent. Once the optimal $\{u_l\}_l$ is obtained, we transform it to $q$ with the softmax function~\eqref{eq:softmax}. 
 
\subsection{Method 2: Matrix factorisation approaches}
\label{sec:method2}
Our second class of approaches is based on low-rank matrix factorisation with missing entries. 
Indeed, it is possible to cast UHC as a problem of filling an incomplete matrix of soft labels.
During the last decade, low-rank matrix completion and factorisation~\cite{Candes2009,DelBue2012} have been successfully used in various applications,~\eg, structure from motion~\cite{Ke2005} and recommender systems~\cite{Koren2009}. It has also been used for multilabel classification in a transductive setting~\cite{Cabral2011}. Here, we will describe how we can use matrix factorisation to recover soft labels $q$ from $\{p_i\}_i$.

\subsubsection{Matrix factorisation in probability space}
\label{sec:MatFactProb}
 Consider a matrix $\mathbf{P}\in[0,1]^{\nclasses\times\nclsfs}$ where we set $P_{li}$ (the element in row $l$ and column $i$) to $p_i(Y=l)$ if $l\in\mathcal{L}_i$ and zero otherwise. This matrix $\mathbf{P}$ is similar to the {\it decision profile matrix} in ensemble methods~\cite{Kuncheva2004}, but here we fill in $0$ for the classes that $C_i$'s cannot predict. To account for these missing predictions, we define $\mathbf{M}\in\{0,1\}^{\nclasses\times\nclsfs}$ as a mask matrix where $\mathbf{M}_{li}$ is 1 if $l\in\mathcal{L}_i$ and zero otherwise. Using the relation between $p_i$ and $q$ in~\eqref{eq:relation_pi_and_q}, we can see that $\mathbf{P}$ can be factorised into a masked product of vectors as:
\begin{equation}
\mathbf{M}\odot\mathbf{P} =
\mathbf{M}
\odot
(\mathbf{u}\mathbf{v}^\top),
\end{equation}
\begin{equation}
	\mathbf{u}=
	\begin{bmatrix}
	q(Y=l_1) \\
	\vdots \\
	q(Y=l_m)
	\end{bmatrix},
	\mathbf{v}=
	\begin{bmatrix}
	\frac{1}{\sum_{l\in\mathcal{L}_1}q(Y=l)} \\
	\vdots \\
	\frac{1}{\sum_{l\in\mathcal{L}_\nclsfs}q(Y=l)}
	\end{bmatrix},
\end{equation}
where $\odot$ is the Hadamard product. Here, $\mathbf{u}$ is the vector containing $q$, and each element in $\mathbf{v}$ contains the normalisation factor for each $C_i$. In this form, we can estimate the probability vector $\mathbf{u}$ by solving the following rank-1 matrix completion problem:
 \begin{align}
 \underset{\mathbf{u},\mathbf{v}}{\minimize} &\ \ \ \Vert\mathbf{M}\odot(\mathbf{P}-\mathbf{u}\mathbf{v}^\top)\Vert_F^2\label{eq:mf_ps}
 \\
 \text{subject to} &\ \ \
  \mathbf{u}^\top\mathbf{1}_{\nclasses}=1
 \\
 &\ \ \
\mathbf{v} \geq \mathbf{0}_{\nclsfs}, \mathbf{u} \geq \mathbf{0}_{\nclasses}  ,
 \end{align}
 where $\Vert\cdot\Vert_F$ denotes Frobenius norm, and $\mathbf{0}_k$ and $\mathbf{1}_k$ denote vectors of zeros and ones of size $k$. Here, the constraints ensure that $\mathbf{u}$ is a probability vector and that $\mathbf{v}$ remains nonnegative so that the sign of probability in $\mathbf{u}$ is not flipped. This formulation can be regarded as a non-negative matrix factorisation problem~\cite{Lee2001}, which we solve using Alternating Least Squares (ALS)~\cite{Berry2007} where we normalise $\mathbf{u}$ to sum to $1$ in each iteration\footnote{We note there are more effective algorithms for matrix factorisation than ALS~\cite{Buchanan2005,Okatani2011,DelBue2012}. Here, we use ALS due to ease of implementation.}. Due to gauge freedom~\cite{Buchanan2005}, this normalisation in $\mathbf{u}$ does not affect the cost function.

\subsubsection{Matrix factorisation in logit space}
\label{sec:MatFactLogit}
In Sec.~\ref{sec:ReviewDistillation}, we discussed the relationship between minimising cross-entropy and logit matching under $\ell_2$ distance. In this section, we consider applying matrix factorisation in logit space and show that our formulation is a generalisation of logit matching between $C_i$ and $C_U$.

Let $z^i_l$ be the given logit output of class $l$ of $C_i$\footnote{For algorithms besides neural networks, we can obtain logits from probability via $z^i_l=\log p_i(Y=l)$.}, and $u_l$ be that of $C_U$ to be estimated.  Consider a matrix $\mathbf{Z}\in\mathbb{R}^{\nclasses\times\nclsfs}$ where $Z_{li}=z^i_l$ if $l\in\mathcal{L}_i$ and zero otherwise. We can formulate the problem of estimating the vector of logits $\mathbf{u}\in\mathbb{R}^\nclasses$ as :
\begin{align}
\underset{\mathbf{u},\mathbf{v},\mathbf{c}}{\minimize} & \ \Vert\mathbf{M}\odot(\mathbf{Z}-\mathbf{u}\mathbf{v}^\top-\mathbf{1}_\nclasses\mathbf{c}^\top )\Vert_F^2+\lambda(\Vert\mathbf{u}\Vert_2^2+\Vert\mathbf{v}\Vert_2^2) 
\label{eq:LowRankLogitOp_full}\\
\text{subject to} & \ \mathbf{v}\geq\mathbf{0}_\nclsfs,
\end{align}
where $\mathbf{c}\in\mathbb{R}^\nclsfs$ deals with shift in logits\footnote{Recall that a shift in logit values has no effect on the probability output, but we need to account for the different shifts from the $C_i$'s to cast it as matrix factorisation.}, and $\lambda\in\mathbb{R}$ is a hyperparameter controlling regularisation~\cite{Buchanan2005}. Here, optimising $\mathbf{v}\in\mathbb{R}^\nclsfs$ is akin to optimising the temperature of logits~\cite{Hinton2015} from each source classifier, and we constrained it to be nonnegative to prevent the logit sign flip, which could affect the probability. 

\textbf{Relation to logit matching} The optimisation in~\eqref{eq:LowRankLogitOp_full} has three variables. Since $\mathbf{c}$ is unconstrained, we derive its closed form solution and remove it from the formulation. This transforms~\eqref{eq:LowRankLogitOp_full} into:
\begin{align}
\underset{\mathbf{u},\mathbf{v}}{\minimize} & \ 
\sum_{i=1}^\nclsfs
\left\Vert
\mathcal{P}_{\vert\mathcal{L}_i\vert}
\left(
[\mathbf{z}_i
-
\mathbf{u}
v_i]_{\mathcal{L}_i}
\right)
\right\Vert_2^2+\lambda(\Vert\mathbf{u}\Vert_2^2+\Vert\mathbf{v}\Vert_2^2) 
\label{eq:LowRankLogitOpt_reduced}
\\
\text{subject to} & \ \mathbf{v}\geq\mathbf{0}_\nclsfs,
\end{align}
where $\mathbf{z}_i$ is the $i^{th}$ column of $\mathbf{Z}$;
 $[\mathbf{x}]_{\mathcal{L}_i}$ selects the elements of $\mathbf{x}$ which are indexed in $\mathcal{L}_i$;
and  $\mathcal{P}_{k}(\mathbf{x})=(\mathbf{I}_{k}-\frac{1}{k}\mathbf{1}_{k}\mathbf{1}_{k}^\top)\mathbf{x}$ is the orthogonal projector that removes the mean from the vector $\mathbf{x}\in\mathbb{R}^k$. This transformation simplifies~\eqref{eq:LowRankLogitOp_full} to contain only $\mathbf{u}$ and $\mathbf{v}$. We can see that this formulation minimises the $\ell_2$ distance between logits, 
but instead of considering all classes in $\mathcal{L}_U$, 
each term in the summation considers only the classes in $\mathcal{L}_i$. In addition,~\eqref{eq:LowRankLogitOpt_reduced} also includes regularisation and optimises for scaling in $\mathbf{v}$. Thus, we can say that~\eqref{eq:LowRankLogitOp_full} is a generalisation of logit matching for UHC.
 
  \textbf{Optimisation} While~\eqref{eq:LowRankLogitOpt_reduced} has fewer parameters than~\eqref{eq:LowRankLogitOp_full}, it is more complicated to optimise as the elements in $\mathbf{u}$ are entangled due to the projector. Instead, we solve~\eqref{eq:LowRankLogitOp_full} using ALS over $\mathbf{u}$, $\mathbf{v}$, and $\mathbf{c}$. Here, there is no constraint on $\mathbf{u}$, so we do not normalise it as in Sec.~\ref{sec:MatFactProb}.
 
 \textbf{Alternative approach: Setting $\mathbf{v}$ as a constant} While setting $\mathbf{v}$ as a variable allows~\eqref{eq:LowRankLogitOp_full} to handle different scalings of logits, it also introduces cumbersome issues. Specifically, the gauge freedom in $\mathbf{u}\mathbf{v}^\top$ may lead to arbitrary scaling in $\mathbf{u}$ and $\mathbf{v}$, \ie, $\mathbf{u}\mathbf{v}^\top=\mathbf{(\mathbf{u}/\alpha)(\alpha\mathbf{v}^\top)}$ for $\alpha\neq0$. Also, while the regularisers help prevent the norms of $\mathbf{u}$  and  $\mathbf{v}$ to become too large, it is difficult to set a single $\lambda$ that works well for all data in $\mathcal{U}$. 
  To combat these issues, we propose another formulation of~\eqref{eq:LowRankLogitOp_full} where we fix $\mathbf{v}=\mathbf{1}_\nclsfs$. With $\mathbf{v}$ fixed, we do not require to regularise $\mathbf{u}$ since its scale is determined by $\mathbf{Z}$. In addition, the new formulation is convex and can be solved to global optimality. We solve this alternative formulation with gradient descent.

 \subsection{Extensions}
 \label{sec:method:directBP}
In Secs.~\ref{sec:method1} and \ref{sec:method2}, we have described methods for estimating $q$ from $\{p_i\}$ then using $q$ as the soft label for training $C_U$. 
In this section, we discuss two possible extensions applicable to all the methods: ({\it i}) direct backpropagation for neural networks and ({\it ii}) fixing imbalance in soft labels.

\subsubsection{Direct backpropagation for neural networks}
\label{sec:method:DirectBP}
Suppose the unified classifier $C_U$ is a neural network. While it possible to use $q$ to train $C_U$ in a supervised manner, we could also consider an alternative where we directly backpropagate the loss without having to estimate  $q$ first. In the case of cross-entropy (Sec.~\ref{sec:method1}), we can think of $q$ as the probability output from $C_U$, through which we can directly backpropagate the loss. In the case of matrix factorisation  (Sec.~\ref{sec:method2}), we could consider $\mathbf{u}$ as the vector of probability (Sec.~\ref{sec:MatFactProb}) or logit (Sec.~\ref{sec:MatFactLogit}) outputs from $C_U$. Once $\mathbf{u}$ is obtained from $C_U$, we plug it in each formulation, solve for other variables (\eg, $\mathbf{v}$ and $\mathbf{c}$) with $\mathbf{u}$ fixed, then backpropagate the loss via $\mathbf{u}$. Directly backpropagating the loss merges the steps of estimating $q$ and using it to train $C_U$ into a single step. 

\subsubsection{Balancing soft labels}
\label{sec:method:BalanceSoftLabels}
All the methods we have discussed are based on individual samples: we estimate $q$ from $\{p_i\}$ of a single $\mathbf{x}$ from the transfer set $\mathcal{U}$ and use it to train $C_U$. However, we observe that the set of estimated $q$'s from the whole $\mathcal{U}$ could be imbalanced. That is, the estimated $q$'s may be biased towards certain classes more than others. To counter this effect, we apply the common technique of weighting the cross-entropy loss while training $C_U$~\cite{Ng2018}. The weight of each class $l$ is computed as the inverse of the mean of $q(Y=l)$ over all data from $\mathcal{U}$.
\section{Experiments}
In this section, we perform experiments to compare different methods for solving UHC. The main experiments on ImageNet, LSUN, and Places365 datasets are described in Sec.~\ref{sec:exp:imagenet_lsun}, while sensitivity analysis is described in Sec.~\ref{sec:exp:ablation}. 

We use the following abbreviations to denote the methods. \textbf{SD} for the naive extension of Standard Distillation (Sec.~\ref{sec:ReviewDistillation})~\cite{Hinton2015}; \textbf{CE-X} for Cross-Entropy methods (Sec.~\ref{sec:method1}); \textbf{MF-P-X} for Matrix Factorization in Probability space (Sec.~\ref{sec:MatFactProb}); and \textbf{MF-LU-X} and \textbf{MF-LF-X} for Matrix Factorization in Logit space with Unfixed and Fixed $\mathbf{v}$ (Sec.~\ref{sec:MatFactLogit}), \textit{resp}. The suffix `\textbf{X}' is replaced with `\textbf{E}' if we estimate $q$ first before using it as soft label to train $C_U$; with `\textbf{BP}' if we perform direct backpropagation from the loss function (Sec.~\ref{sec:method:DirectBP}); and with `{\bf BS}' if we estimate and balance the soft labels $q$ before training $C_U$ (Sec.~\ref{sec:method:BalanceSoftLabels}). In addition to the mentioned methods, we also include \textbf{SD-BS} as the SD method with balanced soft labels, and \textbf{SPV} as the method trained directly in a supervised fashion with all training data of all $C_i$'s as a benchmark. For MF-LU-X methods, we used $\lambda=0.01$. All methods use temperature $T=3$ to smooth the soft labels and logits (See~\eqref{eq:softmax} and~\cite{Hinton2015}).

\subsection{Experiment on large image datasets}
\label{sec:exp:imagenet_lsun}
In this section, we describe our experiment on ImageNet, LSUN, and Places365 datasets. First, we describe the experiment protocols, providing details on the datasets, architectures used as $C_i$ and $C_U$, and the configurations of $C_i$. Then, we discuss the results.

\begin{table}\centering
	\newcommand{\cs}{\hspace{9pt}}
	\renewcommand{\arraystretch}{1}
	\setlength{\aboverulesep}{1.8pt}
	\setlength{\belowrulesep}{1.8pt}
	\caption{HC configurations for the main experiment \label{table:HC_exp_config}}
	{\small
	\begin{tabular}{@{\hspace{2pt}}c@{\cs}c@{\cs}c@{\cs}c@{\cs}c@{\hspace{2pt}}}
		\toprule[1.1px]
		\multirow{2}{*}[-2pt]{Dataset} & \multirow{2}{*}[-1.5pt]{\thead{\small\#Classes \\ \small in $\mathcal{L}_U$ ($L$)}} & \multirow{2}{*}[-2pt]{\#HCs ($N$)} & \multicolumn{2}{c}{\#Classes for each HC}\\
		\cmidrule{4-5} 
		& & & Random & Compl. overlap.\\
		\midrule[1.1px]
		ImageNet & 20-50 & 10-20 & 5-15 & $=L$\\
		LSUN & 5-10 & 3-7 & 2-5 & $=L$ \\
		Places365 & 20-50 & 10-20 & 5-15 & $=L$\\
		\bottomrule[1.1px]
	\end{tabular}
}
\end{table}

\subsubsection{Experiment protocols}
\textbf{Datasets} We use three datasets for this experiment. ({\it i}) ImageNet (ILSVRC2012)~\cite{Russakovsky2015}, consisting of 1k classes with \texttildelow700 to 1300 training and 50 validation images per class, as well as 100k unlabelled test images. In our experiments, the training images are used as training data for the $C_i$'s, the unlabelled test images as $\mathcal{U}$, and the validation images as our test set to evaluate the accuracy. ({\it ii}) LSUN~\cite{Yu2015}, consisting of 10 classes with \texttildelow100k to 3M training and 300 validation images per class with 10k unlabelled test images. Here, we randomly sample a set of 1k training images per class to train the $C_i$'s, a second randomly sampled set of 20k images per class also from the training data is used as $\mathcal{U}$, and the validation data is used as our test set. ({\it iii}) Places365~\cite{Zhou2018}, consisting of 365 classes with \texttildelow3k to 5k training and 100 validation images per class, as well as \texttildelow329k unlabelled test images. We follow the same usage as in ImageNet, but with 100k samples from the unlabelled test images as $\mathcal{U}$. We pre-process all images by centre cropping and scaling to $64\times64$ pixels.

\textbf{HC configurations} We test the proposed methods under two configurations of HCs (see summary in Table~\ref{table:HC_exp_config}). ({\it i}) Random classes. For ImageNet and Places365, in each trial, we sample 20 to 50 classes as $\mathcal{L}_U$ and train 10 to 20 $C_i$'s where each is trained to classify 5 to 15 classes. For LSUN, in each trial, we sample 5 to 10 classes as $\mathcal{L}_U$ and train 3 to 7 $C_i$'s where each is trained to classify 2 to 5 classes. We use this configuration as the main test for when $C_i$'s classify different sets of classes. ({\it ii}) Completely overlapping classes. Here, we use the same configurations as in ({\it i}) except all $C_i$'s are trained to classify all classes in $\mathcal{L}_U$. This case is used to test our proposed methods under the common configurations where all $C_i$ and $C_U$ share the same classes. Under both configurations, $\mathcal{U}$ consist of a much wider set of classes than $\mathcal{L}_U$. In other words, a large portion of the images in $\mathcal{U}$ does not fall under any of the classes in $\mathcal{L}_U$.

\textbf{Models} Each $C_i$ is randomly selected from one of the following four architectures with ImageNet pre-trained weights: AlexNet~\cite{Krizhevsky2012}, VGG16~\cite{Simonyan2014}, ResNet18, and ResNet34~\cite{He2016}. For AlexNet and VGG16, we fix the weights of their feature extractor portion, replace their \texttt{fc} layers with two \texttt{fc} layers with 256 hidden nodes (with BatchNorm and ReLU), and train the \texttt{fc} layers with their training data. Similarly in ResNet models, we replace their \texttt{fc} layers with two \texttt{fc} layers with 256 hidden nodes as above. In addition, we also fine-tune the last residual block. As for $C_U$, we use two models, VGG16 and ResNet34, with similar settings as above. 

For all datasets and configurations, we train each $C_i$ with 50 to 200 samples per class; no sample is shared between any $C_i$ in the same trial. These $C_i$'s together with $\mathcal{U}$ are then used to train $C_U$. We train all models for 20 epochs with SGD optimiser (step sizes of 0.1 and 0.01\footnote{For MF-P-BP, we use $150\times$  the rates as its loss has a smaller scale.} for first and latter 10 epochs with momentum 0.9). To control the variation in results, in each trial we initialise instances of $C_U$'s from the same architecture using the same weights and we train them using the same batch order. In each trial, we evaluate the $C_U$'s of all methods on the test data from all classes in $\mathcal{L}_U$. We run 50 trials for each dataset, model, and HC configuration combination. The results are reported in the next section.

\begin{table*}
	\caption{Average accuracy of UHC methods over different combinations of HC configurations, datasets, and unified classifier models. (\underline{\bf Underline bold}: Best method. {\bf Bold}: Methods which are not statistically significantly different from the best method.)\label{table:results}}
	{
		\definecolor{gray}{rgb}{0.85,0.85,0.85}
		\newcommand{\hs}{\hspace{4pt}}
		\newcommand{\uline}{\underline}
		\footnotesize
		\renewcommand{\arraystretch}{1}
		\setlength{\aboverulesep}{1.5pt}
		\setlength{\belowrulesep}{1.5pt}
		\begin{tabular}{@{}c@{}c@{}c@{\hs}c@{\hs}c@{\hs}c@{\hs}c@{\hs}c@{\hs}c@{\hs}c@{\hs}c@{\hs}c@{\hs}c@{\hs}c@{\hs}c@{\hs}c@{\hs}c@{\hs}c@{\hs}c@{\hs}}
			\toprule[1.1px]
			\multicolumn{2}{c}{} & \multicolumn{8}{c}{Random Classes} &  & \multicolumn{8}{c}{Completely Overlapping Classes}\tabularnewline
			\cmidrule{3-10} \cmidrule{12-19}
			\multicolumn{2}{c}{Methods} & \multicolumn{2}{c}{ImageNet} &  & \multicolumn{2}{c}{LSUN} &  & \multicolumn{2}{c}{Places365} &  & \multicolumn{2}{c}{ImageNet} &  & \multicolumn{2}{c}{LSUN} &  & \multicolumn{2}{c}{Places365}\tabularnewline
			\cmidrule(r){3-4} \cmidrule(r){6-7} \cmidrule(r){9-10} \cmidrule(r){12-13} \cmidrule(r){15-16} \cmidrule{18-19}
			\multicolumn{2}{c}{} & VGG16 & ResNet34 &  & VGG16 & ResNet34 &  & VGG16 & ResNet34 &  & VGG16 & ResNet34 &  & VGG16 & ResNet34 &  & VGG16 & ResNet34\tabularnewline
			\midrule[1.1px]
			\multicolumn{2}{r}{SPV (Benchmark)} & .7212 & .6953 &  & .6664 & .6760 &  & .5525 & .5870 &  & .7345 & .7490 &  & .6769 & .7017 &  & .5960 & .6460\tabularnewline
			\cmidrule{1-19}
			\multicolumn{2}{r}{SD} & .5543 & .5562 &  & .5310 & .5350 &  & .4390 & .4564 &  & \textbf{.7275} & \textbf{.7292} &  & .7004 & \textbf{.7041} &  & \textbf{.6163} & \textbf{.6402}\tabularnewline
			\cmidrule{1-19}
			\multicolumn{2}{l}{\textbf{(A)} {\it Estimate $q$ methods}} &  &  &  &  &  &  &  &  &  &  &  &  &  &  &  &  & \tabularnewline
			\multicolumn{2}{r}{CE-E} & \textbf{.6911} & \textbf{.6852} &  & .6483 & \textbf{.6445} &  & \textbf{.5484} & \textbf{.5643} &  & \textbf{.7276} & \textbf{.7290} &  & .7002 & .7036 &  & \textbf{.6162} & \underline{\bf .6406}\tabularnewline
			\multicolumn{2}{r}{MF-P-E} & .6819 & .6747 &  & .6443 & .6406 &  & .5349 & .5488 &  & \textbf{\uline{.7280}} & \textbf{\uline{.7297}} &  & \textbf{.7012} & \textbf{.7052} &  & \textbf{\uline{.6167}} & \underline{\bf .6406}\tabularnewline
			\multicolumn{2}{r}{MF-LV-E} & .6660 & .6609 &  & .6348 & .6330 &  & .5199 & .5414 &  & .7231 & .7242 &  & \textbf{\uline{.7031}} & \textbf{.7043} &  & .6129 & .6374\tabularnewline
			\multicolumn{2}{r}{MF-LF-E} & .6886 & \textbf{.6833} &  & .6490 & \textbf{.6458} &  & .5441 & .5609 &  & .7265 & \textbf{.7279} &  & \textbf{.7015} & \textbf{\uline{.7057}} &  & \textbf{.6161} & \textbf{.6397}\tabularnewline
			\cmidrule{1-19}
			\multicolumn{2}{l}{\textbf{(B)} {\it Backprop. methods}} &  &  &  &  &  &  &  &  &  &  &  &  &  &  &  &  & \tabularnewline
			\multicolumn{2}{r}{CE-BP} & \textbf{.6902} & \textbf{.6869} &  & \textbf{.6520} & .6439 &  & .5466 & \textbf{.5669} &  & \textbf{.7275} & \textbf{.7288} &  & .7003 & \textbf{.7040} &  & \textbf{.6161} & \textbf{.6400}\tabularnewline
			\multicolumn{2}{r}{MF-P-BP} & \textbf{\uline{.6945}} & \textbf{\uline{.6872}} &  & .6480 & \textbf{.6417} &  & \textbf{.5471} & .5609 &  & \textbf{.7277} & \textbf{.7287} &  & .6999 & \textbf{.7019} &  & .6146 & .6384\tabularnewline
			\multicolumn{2}{r}{MF-LV-BP} & .6889 & \textbf{.6847} &  & \textbf{.6495} & .6389 &  & .5467 & \textbf{.5681} &  & .7229 & .7225 &  & .7001 & \textbf{.7046} &  & .6113 & .6369\tabularnewline
			\multicolumn{2}{r}{MF-LF-BP} & .6842 & \textbf{.6840} &  & \textbf{.6523} & \textbf{.6445} &  & .5383 & \textbf{.5624} &  & .7239 & .7252 &  & \textbf{.7020} & .7034 &  & .6104 & .6366\tabularnewline
			\cmidrule{1-19}
			\multicolumn{2}{l}{\textbf{(C)} {\it Balanced soft labels}} &  &  &  &  &  &  &  &  &  &  &  &  &  &  &  &  & \tabularnewline
			\multicolumn{2}{r}{SD-BS} & .6629 & .6574 &  & .6343 & .6345 &  & .5283 & .5433 &  & .7217 & .7214 &  & .6979 & .7017 &  & .6094 & .6320\tabularnewline
			\multicolumn{2}{r}{CE-BS} & \textbf{.6928} & \textbf{.6856} &  & .6513 & \textbf{.6464} &  & \textbf{\uline{.5548}} & \textbf{.5687} &  & .7215 & .7213 &  & .6979 & .7018 &  & .6094 & .6323\tabularnewline
			\multicolumn{2}{r}{MF-P-BS} & .6851 & .6756 &  & .6474 & \textbf{.6450} &  & .5455 & .5546 &  & .7243 & .7252 &  & .6996 & .7041 &  & .6124 & .6355\tabularnewline
			\multicolumn{2}{r}{MF-LV-BS} & .6772 & .6682 &  & .6388 & .6357 &  & .5346 & .5497 &  & .7168 & .7173 &  & .7014 & .7028 &  & .6063 & .6301\tabularnewline
			\multicolumn{2}{r}{MF-LF-BS} & \textbf{.6935} & \textbf{.6865} &  & \textbf{\uline{.6549}} & \textbf{\uline{.6485}} &  & \textbf{.5544} & \textbf{\uline{.5692}} &  & .7210 & .7215 &  & .6998 & .7035 &  & .6101 & .6330\tabularnewline
			\bottomrule[1.1px]
		\end{tabular}
	}
\end{table*}

\subsubsection{Results}

Table~\ref{table:results} shows the results for this experiment. Each column shows the average accuracy of each method under each experiment setting, where the best performing method is shown in underlined bold. To test statistical significance, we choose Wilcoxon signed-rank test over standard deviation to cater for the vastly different settings (\eg, model architectures, number of classes and HCs, \etc.) across trials. We run the test between the best performing method in each experiment and the rest. Methods where the performance is not statistically significantly different from the best method at $\alpha=0.01$ are shown in bold. 

First, let us observe the result for the {\it random classes} case which addresses the main scenario of this paper, \ie, when each HC is trained to classify different sets of classes. We can make the following observations.

\textbf{All proposed methods perform significantly better than SD.} We can see that all methods in (A), (B), and (C) of Table~\ref{table:results} outperform SD by a large margin of 9-15\%. This shows that simply setting probability of undefined classes in each HC to $0$ may significantly deteriorate the accuracy. On the other hand, our proposed methods achieve significantly better results and almost reach the same accuracy as SPV with a gap of 1-4\%. This suggests the soft labels from HCs can be used for unsupervised training at a little expense of accuracy, even though $\mathcal{U}$ contains a significant proportion of images that are not part of the target classes. Still, there are several factors that may affect the capability of $C_U$ from reaching the accuracy of SPV, \eg, accuracy of $C_i$, their architectures, \etc. We look at some of these in the sensitivity analysis section.

\textbf{MF-LF-BS performs well in all cases.} We can see that different algorithms perform best under different settings, but MF-LF-BS always performs best or has no statistical difference from the best methods. This suggests MF-LF-BS could be the best method for solving UHC. At the same time, CE methods offer a good trade-off between high accuracy and ease of implementation, which makes them a good alternative for the UHC problem.

Besides these main points, we also note the following small but consistent trends.

\textbf{Balancing soft labels helps improve accuracy.} While the improvement may be marginal (less than 1.5\%), we can see that `BS' methods in (C) consistently outperform their `E' counterparts in (A). Surprisingly, SD-BS, which is SD with balanced soft labels, also significantly improved over SD by more than 10\%. These results indicate that it is a good practice to use balanced soft labels to solve UHC. Note that while SD-BS received significant boost, it still generally underperforms compared to CE and MF methods, suggesting that it is important to incorporate the relation between $\{p_i\}$ and $q$ into training.

\textbf{Nonconvex losses perform better with `BP'.} Methods with suffixes `E' and `BS' in (A) and (C) are based on estimating $q$ before training $C_U$, while `BP' in (B) directly perform backpropagation from the loss function. As seen in Sec.~\ref{sec:method}, the losses of CE and MF-LF are convex in their variables while MF-P and MF-LV are nonconvex. Here, we observe a small but interesting effect that methods with nonconvex losses perform better with `BP'. We speculate that this is due to errors in the estimation of $q$ trickling down to the training of $C_U$ if the two steps are separated. Conversely in `BP', where the two steps are merged into a single step, such issue might be avoided. More research would be needed to confirm this speculation. For convex losses (CE and MF-LF), we find no significant patterns between `E' in (A) and `BP' in (B). 

\begin{figure*}
	\centering
	\includegraphics[width=1\linewidth]{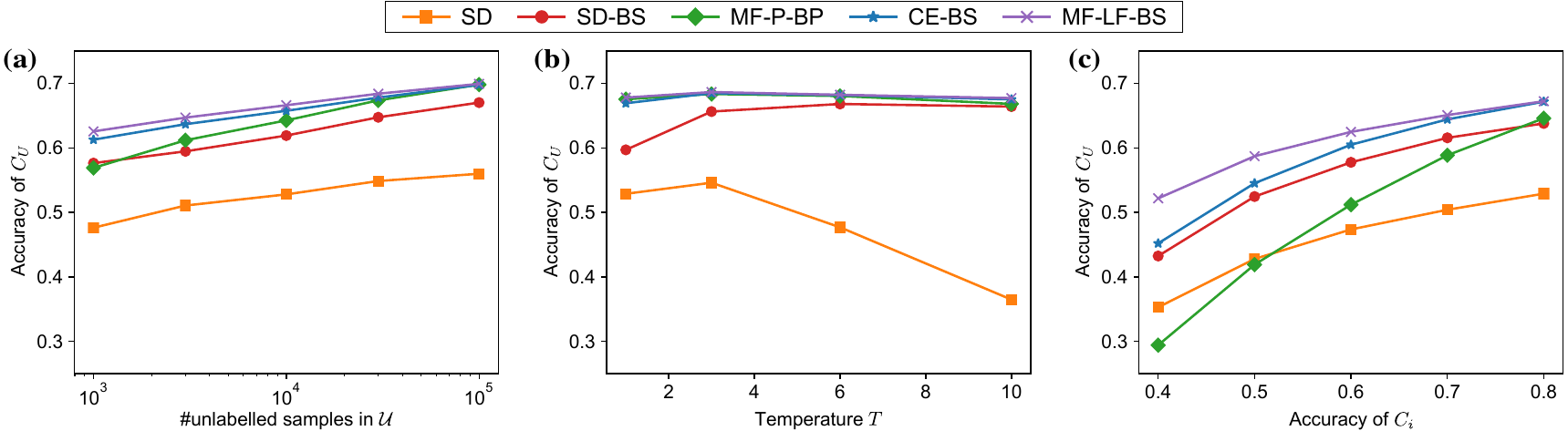}
	\caption{Sensitivity analysis results. (a) Size of unlabelled set. (b) Temperature. (c) Accuracy of HCs.}
	\label{fig:ablation}	
\end{figure*}

Next, we discuss the {\it completely overlapping case}.

\textbf{All methods perform rather well.} We can see that all methods, including SD, achieve about the same accuracy (within \texttildelow1\% range). This shows that our proposed methods can also perform well in the common cases of all $C_i$'s being trained to classify all classes and corroborates the claim that our proposed methods are generalisations of distillation.

\textbf{Not balancing soft labels performs better.} We note that balancing soft labels tends to slightly deteriorate the accuracy. This is the opposite result from the random classes case. Here, even SD-BS which receive an accuracy boost in the random classes case also performs worse than its counterpart SD. This suggests not balancing soft labels may be a better option for overlapping classes case.

\textbf{Distillation may outperform its supervised counterparts.} For LSUN and Places365 datasets, we see that many times distillation methods performs better than SPV. Especially for the case of VGG16, we see SPV consistently perform worse than other methods by 1 to 3\%  in most of the trials. This shows that it is possible that distillation-based methods may outperform their supervised counterparts.

\subsection{Sensitivity Analysis}
\label{sec:exp:ablation}
In this section, we perform three sets of sensitivity analysis on the effect of size of the transfer set, temperature parameter $T$, and accuracy of HCs. We use the same settings as the ImageNet random classes experiment in the previous section with VGG16 as $C_U$. We run 50 trials for each test. We evaluate the following five methods as the representative set of SD and top performing methods from previous section: SD, SD-BS, MF-P-BP, MF-LF-BS, and CE-BS. 

\textbf{Size of transfer set} We use this test to evaluate the effect of the number of unlabelled samples in the transfer set $\mathcal{U}$. We vary the number of samples from $10^3$ to $10^5$. The result is shown in Fig.~\ref{fig:ablation}a. As expected, we can see that all methods deteriorate as the size of transfer set decreases. In this test, MF-P-BP is the most affected by the decrease as its accuracy drops fastest. Still, all other methods perform better than SD in the whole test range, illustrating the robustness to transfer sets with different sizes.

\textbf{Temperature} In this test, we vary the temperature $T$ used for smoothing the probability $\{p_i\}$ (see~\eqref{eq:softmax} or~\cite{Hinton2015}) before using them to estimate $q$ or train $C_U$. The values evaluated are $T=1, 3, 6,  \text{and}\ 10$. The result is shown in Fig.~\ref{fig:ablation}b. We can see that the accuracies of SD and SD-BS drop significantly when $T$ is set to high and low values, {\it resp}. On the other hand, the other three methods are less affected by different values of $T$.

\textbf{HCs' accuracies} In this test, we evaluate the robustness of UHC methods against varying accuracy of $C_i$. The test protocol is as follows. In each trial, we vary the accuracy of all $C_i$'s to 40-80\%, obtain ${p_i}$ from the $C_i$'s, and use them to perform UHC. To vary the accuracy of each $C_i$, we take $50$ samples per class from training data as the adjustment set, completely train each $C_i$ from the remaining training data, then inject increasing Gaussian noise into the last \texttt{fc} layer until its accuracy on the adjustment set drops to the desired value. If the initial accuracy of $C_i$ is below the desired value then we simply use the initial $C_i$. The result of this evaluation is shown in Fig.~\ref{fig:ablation}c. We can see that the accuracy of all methods increase as the $C_i$'s perform better, illustrating that the accuracy of $C_i$ is an important factor for the performance of UHC methods. We can also see that MF-P-BP is most affected by low accuracy of $C_i$ while MF-LF-BS is the most robust. 

Based on the sensitivity analysis, we see that MF-LF-BS is the most robust method against the number of samples in the transfer set, temperature, and accuracy of the HCs. This result provides further evidence that MF-LF-BS should be the suggested method for solving UHC. We provide the complete sensitivity plots with all methods in the supplementary material.

\section{Conclusion}
In this paper, we formalise the problem of unifying knowledge from heterogeneous classifiers (HCs) using only unlabelled data. We proposed cross-entropy minimisation and matrix factorisation methods for estimating soft labels of the unlabelled data from the output of HCs based on a derived probabilistic relationship. We also proposed two extensions to directly backpropagate the loss for neural networks and to balance estimated soft labels. Our extensive experiments on ImageNet, LSUN, and Places365 show that our proposed methods significantly outperformed a naive extension of knowledge distillation. The result together with additional three sensitivity analysis suggest that an approach based on matrix factorization in logit space with balanced soft labels is the most robust approach to unify HCs into a single classfier.

{\small
	\bibliographystyle{ieee}
	\bibliography{egbib_uhc}

\begin{thebibliography}{10}\itemsep=-1pt

\bibitem{Avidan2007}
Shai Avidan.
\newblock Ensemble tracking.
\newblock {\em IEEE TPAMI}, 29(2):261--271, 2007.

\bibitem{Berry2007}
Michael~W. Berry, Murray Browne, Amy~N. Langville, Paul~V. Pauca, and Robert~J.
  Plemmons.
\newblock Algorithms and applications for approximate nonnegative matrix
  factorization.
\newblock {\em Computational statistics \& data analysis}, 52(1):155--173,
  2007.

\bibitem{Boyd2007}
Stephen Boyd, Seung-Jean Kim, Lieven Vandenberghe, and Arash Hassibi.
\newblock A tutorial on geometric programming.
\newblock {\em Optimization and engineering}, 8(1):67, 2007.

\bibitem{Boyd2011}
Stephen Boyd, Neal Parikh, Eric Chu, Borja Peleato, and Jonathan Eckstein.
\newblock Distributed optimization and statistical learning via the alternating
  direction method of multipliers.
\newblock {\em Foundations and Trends in Machine Learning}, 3(1):1--122, Jan.
  2011.

\bibitem{Boyd2004}
Stephen Boyd and Lieven Vandenberghe.
\newblock {\em Convex optimization}.
\newblock Cambridge university press, 2004.

\bibitem{Breiman2001}
Leo Breiman.
\newblock Random forests.
\newblock {\em Machine learning}, 45(1):5--32, 2001.

\bibitem{Buchanan2005}
Aeron~M. Buchanan and Andrew~W. Fitzgibbon.
\newblock Damped newton algorithms for matrix factorization with missing data.
\newblock In {\em CVPR}, 2005.

\bibitem{Bucila2006}
Cristian Bucila, Rich Caruana, and Alexandru Niculescu{-}Mizil.
\newblock Model compression.
\newblock In {\em {ACM} {SIGKDD}}, pages 535--541, 2006.

\bibitem{Cabral2011}
Ricardo~S. Cabral, Fernando De~la Torre, Jo{\~a}o~P. Costeira, and Alexandre
  Bernardino.
\newblock Matrix completion for multi-label image classification.
\newblock In {\em NIPS}, 2011.

\bibitem{Candes2009}
Emmanuel~J. Cand{\`e}s and Benjamin Recht.
\newblock Exact matrix completion via convex optimization.
\newblock {\em Foundations of Computational mathematics}, 9(6):717, 2009.

\bibitem{DelBue2012}
Alessio Del~Bue, Joao Xavier, Lourdes Agapito, and Marco Paladini.
\newblock Bilinear modeling via augmented lagrange multipliers ({BALM}).
\newblock {\em IEEE TPAMI}, 34(8):1496--1508, 2012.

\bibitem{Forero2010}
Pedro~A. Forero, Alfonso Cano, and Georgios~B. Giannakis.
\newblock Consensus-based distributed support vector machines.
\newblock {\em JMLR}, 11:1663--1707, Aug. 2010.

\bibitem{Freund1999}
Yoav Freund and Robert Schapire.
\newblock A short introduction to boosting.
\newblock {\em Journal-Japanese Society For Artificial Intelligence},
  14(771-780):1612, 1999.

\bibitem{Gupta2016}
Saurabh Gupta, Judy Hoffman, and Jitendra Malik.
\newblock Cross modal distillation for supervision transfer.
\newblock In {\em CVPR}, 2016.

\bibitem{Hastie2009}
Trevor Hastie, Saharon Rosset, Ji Zhu, and Hui Zou.
\newblock Multi-class adaboost.
\newblock {\em Statistics and its Interface}, 2(3):349--360, 2009.

\bibitem{He2016}
Kaiming He, Xiangyu Zhang, Shaoqing Ren, and Jian Sun.
\newblock Deep residual learning for image recognition.
\newblock In {\em CVPR}, 2016.

\bibitem{Hinton2015}
Geoffrey~E. Hinton, Oriol Vinyals, and Jeffrey Dean.
\newblock Distilling the knowledge in a neural network.
\newblock In {\em {NIPS} Deep Learning and Representation Learning Workshop},
  2015.

\bibitem{Ke2005}
Qifa Ke and Takeo Kanade.
\newblock Robust l1 norm factorization in the presence of outliers and missing
  data by alternative convex programming.
\newblock In {\em CVPR}, 2005.

\bibitem{Kittler1998}
Josef Kittler, Mohamad Hatef, Robert P.~W. Duin, and Jiri Matas.
\newblock On combining classifiers.
\newblock {\em IEEE TPAMI}, 20(3):226--239, 1998.

\bibitem{Konecny2016}
Jakub Konečný, Brendan~H. McMahan, Felix~X. Yu, Peter Richtarik,
  Ananda~Theertha Suresh, and Dave Bacon.
\newblock Federated learning: Strategies for improving communication
  efficiency.
\newblock In {\em NIPS Workshop on Private Multi-Party Machine Learning}, 2016.

\bibitem{Koren2009}
Yehuda Koren, Robert Bell, and Chris Volinsky.
\newblock Matrix factorization techniques for recommender systems.
\newblock {\em Computer}, (8):30--37, 2009.

\bibitem{Krizhevsky2012}
Alex Krizhevsky, Ilya Sutskever, and Geoffrey~E. Hinton.
\newblock Imagenet classification with deep convolutional neural networks.
\newblock In {\em NIPS}, 2012.

\bibitem{Kuncheva2004}
Ludmila~I. Kuncheva.
\newblock {\em Combining pattern classifiers: methods and algorithms}.
\newblock John Wiley \& Sons, 2004.

\bibitem{Lee2001}
Daniel~D Lee and Sebastian~H. Seung.
\newblock Algorithms for non-negative matrix factorization.
\newblock In {\em NIPS}, 2001.

\bibitem{Lin2014}
Tsung-Yi Lin, Michael Maire, Serge Belongie, James Hays, Pietro Perona, Deva
  Ramanan, Piotr Doll{\'a}r, and C~Lawrence Zitnick.
\newblock Microsoft coco: Common objects in context.
\newblock In {\em ECCV}, 2014.

\bibitem{Lopes2017}
Raphael~Gontijo Lopes, Stefano Fenu, and Thad Starner.
\newblock Data-free knowledge distillation for deep neural networks.
\newblock In {\em {NIPS} workshop on learning with limited labeled data}, 2017.

\bibitem{Malisiewicz2011}
Tomasz Malisiewicz, Abhinav Gupta, and Alexei~A Efros.
\newblock Ensemble of exemplar-{SVM}s for object detection and beyond.
\newblock In {\em ICCV}, 2011.

\bibitem{Ng2018}
Andrew Ng.
\newblock {\em Machine Learning Yearning}, chapter~39, page~76.
\newblock deeplearning.ai, 2018.

\bibitem{Okatani2011}
Takayuki Okatani, Takahiro Yoshida, and Koichiro Deguchi.
\newblock Efficient algorithm for low-rank matrix factorization with missing
  components and performance comparison of latest algorithms.
\newblock In {\em ICCV}, 2011.

\bibitem{Polikar2006}
Robi Polikar.
\newblock Ensemble based systems in decision making.
\newblock {\em IEEE Circuits and Systems Magazine}, 6(3):21--45.

\bibitem{Romero2015}
Adriana Romero, Nicolas Ballas, Samira~Ebrahimi Kahou, Antoine Chassang, Carlo
  Gatta, and Yoshua Bengio.
\newblock Fit{N}ets: Hints for thin deep nets.
\newblock In {\em {ICLR}}, 2015.

\bibitem{Russakovsky2015}
Olga Russakovsky, Jia Deng, Hao Su, Jonathan Krause, Sanjeev Satheesh, Sean Ma,
  Zhiheng Huang, Andrej Karpathy, Aditya Khosla, Michael Bernstein, et~al.
\newblock Imagenet large scale visual recognition challenge.
\newblock {\em IJCV}, 115(3):211--252, 2015.

\bibitem{Simonyan2014}
Karen Simonyan and Andrew Zisserman.
\newblock Very deep convolutional networks for large-scale image recognition.
\newblock {\em arXiv preprint arXiv:1409.1556}, 2014.

\bibitem{Viola2001}
Paul Viola and Michael Jones.
\newblock Rapid object detection using a boosted cascade of simple features.
\newblock In {\em CVPR}, 2001.

\bibitem{Xu2017}
Zheng Xu, Yen-Chang Hsu, and Jiawei Huang.
\newblock Learning loss for knowledge distillation with conditional adversarial
  networks.
\newblock In {\em {ICLR} workshop}, 2017.

\bibitem{Yu2015}
Fisher Yu, Ari Seff, Yinda Zhang, Shuran Song, Thomas Funkhouser, and Jianxiong
  Xiao.
\newblock {LSUN}: Construction of a large-scale image dataset using deep
  learning with humans in the loop.
\newblock {\em arXiv preprint arXiv:1506.03365}, 2015.

\bibitem{Zhou2018}
Bolei Zhou, Agata Lapedriza, Aditya Khosla, Aude Oliva, and Antonio Torralba.
\newblock Places: A 10 million image database for scene recognition.
\newblock {\em IEEE transactions on pattern analysis and machine intelligence},
  40(6):1452--1464, 2018.

\end{thebibliography}


\begin{thebibliography}{1}\itemsep=-1pt

\bibitem{Berry2007}
Michael~W. Berry, Murray Browne, Amy~N. Langville, Paul~V. Pauca, and Robert~J.
  Plemmons.
\newblock Algorithms and applications for approximate nonnegative matrix
  factorization.
\newblock {\em Computational statistics \& data analysis}, 52(1):155--173,
  2007.

\bibitem{Boyd2007}
Stephen Boyd, Seung-Jean Kim, Lieven Vandenberghe, and Arash Hassibi.
\newblock A tutorial on geometric programming.
\newblock {\em Optimization and engineering}, 8(1):67, 2007.

\end{thebibliography}
}

\clearpage
\appendix
\twocolumn[\section*{\centering \Large Supplementary Material}]

\section*{Contents}
This supplementary material contains the following contents.
\begin{itemize}
	\item Sec.~\ref{sec:CEGP} Cross-Entropy Method and Geometric Program
	\item Sec.~\ref{sec:ALS} Alternating Least Squares (ALS) for Matrix Factorisation Methods
	\item Sec.~\ref{sec:CompCost} Computation cost
	\item Sec.~\ref{sec:SenAna} Complete results for sensitivity analysis
\end{itemize}

\section{Cross-Entropy Method and Geometric Program}
\label{sec:CEGP}
In this section, we show how to transform~\eqref{eq:CrossEntropy_Unify} in the main paper into a geometric program~\cite{Boyd2007}. First, we rewrite $J(q)$ as follows:
\begin{align}
J(q) &= -\sum_i \sum_{l\in\mathcal{L}_i} p_i(X=l) \log \frac{q(X=l)}{\sum_{k\in\mathcal{L}_i}q(X=k)} \label{eq:CE_supp}
\\
&= \log \frac
{\prod_i\prod_{l\in\mathcal{L}_i}\left(\sum_{k\in\mathcal{L}_i}q(X=k)\right)^{p_i(X=l)}}
{\prod_i\prod_{l\in\mathcal{L}_i}q(X=l)^{p_i(X=l)}}.
\end{align}
Then, we can transform the following problem
\begin{equation}
\underset{q}{\minimize} \ \ J(q)=\log \frac
{\prod_i\prod_{l\in\mathcal{L}_i}\left(\sum_{k\in\mathcal{L}_i}q(X=k)\right)^{p_i(X=l)}}
{\prod_i\prod_{l\in\mathcal{L}_i}q(X=l)^{p_i(X=l)}}
\end{equation}
into 
\begin{align}
\underset{q, \{t_i\}_i}{\minimize} & \ \ \log \frac
{\prod_i\prod_{l\in\mathcal{L}_i}t_i^{p_i(X=l)}}
{\prod_i\prod_{l\in\mathcal{L}_i}q(X=l)^{p_i(X=l)}} \label{eq:geometricProgramCE}
\\
\text{subject to} 
&
\ \ \sum_{k\in\mathcal{L}_i}q(X=k) \leq t_i, i=1,\dots,N,
\end{align}
where we add new variables $t_i, i=1,\dots,N,$ to upperbound each posynomial term in the numerator of the objective function. This turns the objective into a $\log$ of a monomial and adds inequality constraints to the formulation. Since $\log$ is an increasing function and its argument in the objective function is a monomial, removing $\log$ from the objective does not affect the minimum. This leads us to 
\begin{align}
\underset{q, \{t_i\}_i}{\minimize} & \ \ \frac
{\prod_i\prod_{l\in\mathcal{L}_i}t_i^{p_i(X=l)}}
{\prod_i\prod_{l\in\mathcal{L}_i}q(X=l)^{p_i(X=l)}} \label{eq:geometricProgramCE}
\\
\text{subject to} 
&
\ \ \sum_{k\in\mathcal{L}_i}q(X=k) \leq t_i, i=1,\dots,N.
\end{align}
which is a geometric program with variables $q$ and $t_i$~\cite{Boyd2007}. With this formulation, we can further transform it into a convex problem with a change of variable. Here, we define $u_l\in\mathbb{R}$ for $l\in\mathcal{L}_U$ as $q(X=l) = \exp(u_l)$ (\ie, $u_l=\log q(X=l)$). Instead of changing $q$ in~\eqref{eq:geometricProgramCE}, we directly change $q$ in~\eqref{eq:CE_supp}. This transforms $J(q)$ to 
\begin{equation}
\hat{J}(\{u_l\}_l) = -\sum_i \sum_{l\in\mathcal{L}_i} p_i(X=l) \left(
u_l-\log
\left(
\sum_{k\in\mathcal{L}_i}\exp(u_k)
\right)
\right),
\end{equation}
which is~\eqref{eq:CrossEntropy_Unify_Optimise} in the main paper.

\section{Alternating Least Squares (ALS) for Matrix Factorisation Methods}
\label{sec:ALS}
In this section, we detail the Alternative Least Squares (ALS)~\cite{Berry2007} algorithms used for matrix factorisation in the main paper. 

\subsection{ALS for matrix factorisation in probability space}
\label{sec:ALS_MF_prob}
First, let us recall the formulation (\eqref{eq:mf_ps} in the main paper):

 \begin{align}
 \underset{\mathbf{u},\mathbf{v}}{\minimize} &\ \ \ \Vert\mathbf{M}\odot(\mathbf{P}-\mathbf{u}\mathbf{v}^\top)\Vert_F^2
 \\
 \text{subject to} &\ \ \
 \mathbf{u}^\top\mathbf{1}_{\nclasses}=1 \label{eq:MFProbSumConst}
 \\
 &\ \ \
 \mathbf{v} \geq \mathbf{0}_{\nclsfs}, \mathbf{u} \geq \mathbf{0}_{\nclasses}  \label{eq:MFProbNonNegConst},
 \end{align}
The ALS algorithm for solving the above formulation is shown in Alg.~\ref{alg:solveMFinProb}. Steps 4 and 12 are derived from the closed-form solution of $\mathbf{u}$ and $\mathbf{v}$ in the cost function, {\it resp}. Steps 5 and 13 project $\mathbf{u}$ and $\mathbf{v}$ to the nonnegative orthants to satisfy the constraints in \eqref{eq:MFProbNonNegConst}. Steps 7 to 10 are for normalising $\mathbf{u}$ to sum to $1$ per constraint~\eqref{eq:MFProbSumConst}. In fact, for this algorithm, steps 5 and 13 are actually not necessary. This is because all $u_j$'s from step 4 and $v_i$'s from step 12 are already nonnegative since they are the results of division between nonnagative numbers. For termination criteria, we terminate the algorithm if the RMSE between different iterations of $\mathbf{u}$ and $\mathbf{v}$ is less than $10^{-3}$. We also use the maximum number of iterations of $3000$ as a termination criteria.

In terms of implementation, each for-loop can be computed with vector operations (\eg, in MATLAB or with Numpy in Python) instead of using for-loops. In addition, the factorisation of different samples can be performed in parallel on GPUs. These techniques allow a significant speed up compared with the naive implementation.

\begin{algorithm}
	\caption{Matrix factorisation in probabilty space\label{alg:solveMFinProb}}
	\begin{algorithmic}[1]
		\REQUIRE $\mathbf{M}$, $\mathbf{P}$
		\ENSURE $\mathbf{u}$, $\mathbf{v}$
		\STATE Initialise $\mathbf{v}\coloneqq\mathbf{1}_\nclsfs$
		\WHILE {not converged}
		\FOR{$j\coloneqq1, \dots, \nclasses$}
		\STATE \small $u_j\coloneqq\left(\sum_{i=1}^\nclsfs M_{ji}P_{ji}v_i\right)
		/
		\left(\sum_{i=1}^\nclsfs M_{ji}v_i^2\right)$
		\STATE $u_j\coloneqq\max(0,u_j)$
		\ENDFOR
		\STATE $\bar{u}\coloneqq\sum_{j=1}^\nclasses u_j$
		\FOR {$j\coloneqq1, \dots, \nclasses$}
		\STATE $u_j\coloneqq u_j/\bar{u}$
		\ENDFOR
		\FOR {$i\coloneqq1, \dots, \nclsfs$}
		\STATE \small $v_i\coloneqq\left(\sum_{j=1}^\nclasses M_{ji}P_{ji}u_j\right)
		/
		\left(\sum_{j=1}^\nclasses M_{ji}u_j^2\right)$
		\STATE $v_i\coloneqq\max(0,v_i)$
		\ENDFOR
		\ENDWHILE
	\end{algorithmic}
\end{algorithm}

\subsection{ALS for matrix factorisation in logit space}
Again, let us recall the formulation (\eqref{eq:LowRankLogitOp_full} in the main paper):

\begin{align}
\underset{\mathbf{u},\mathbf{v},\mathbf{c}}{\minimize} & \ \Vert\mathbf{M}\odot(\mathbf{Z}-\mathbf{u}\mathbf{v}^\top-\mathbf{1}_\nclasses\mathbf{c}^\top )\Vert_F^2+\lambda(\Vert\mathbf{u}\Vert_2^2+\Vert\mathbf{v}\Vert_2^2) 
\\
\text{subject to} & \ \mathbf{v}\geq\mathbf{0}_\nclsfs,
\end{align}
The ALS for solving the above formulation is shown in Alg.~\ref{alg:solveMFinLogit}. The derivation is similar to that in Sec.~\ref{sec:ALS_MF_prob}. That is, each step is derived via the closed-form solution of each variable, followed by appropriate projection steps. We use the same termination criteria as in previous section.

\begin{algorithm}
	\caption{Matrix factorisation in logit space\label{alg:solveMFinLogit}}
	\begin{algorithmic}[1]
		\REQUIRE $\mathbf{M}$, $\mathbf{Z}$, $\lambda$
		\ENSURE $\mathbf{u}$, $\mathbf{v}$, $\mathbf{c}$
		\STATE Initialise $c_i\coloneqq\left(\sum_{j=1}^\nclasses M_{ji}Z_{ji}\right)/\left(\sum_{j=1}^\nclasses M_{ji}\right),\forall i$
		\STATE Initialise $\mathbf{v}\coloneqq\mathbf{1}_\nclsfs$
		\WHILE {not converged}
		\FOR {$j\coloneqq1, \dots, \nclasses$}
			\STATE \small $u_j\coloneqq\left(\sum_{i=1}^\nclsfs M_{ji}(Z_{ji}-c_i)v_i\right)
			/
			\left(\lambda+\sum_{i=1}^\nclsfs M_{ji}v_i^2\right)$
		\ENDFOR
		\FOR {$i\coloneqq1, \dots, \nclsfs$}
			\STATE \small $v_i\coloneqq\left(\sum_{j=1}^\nclasses M_{ji}(Z_{ji}-c_i)u_j\right)
			/
			\left(\lambda+\sum_{j=1}^\nclasses M_{ji}u_j^2\right)$
			\STATE $v_i\coloneqq\max(0,v_i)$
		\ENDFOR
		\FOR {$i\coloneqq1, \dots, \nclsfs$}
			\STATE $c_i\coloneqq\left(\sum_{j=1}^\nclasses M_{ji}(Z_{ji}-u_jv_i)\right)/\left(\sum_{j=1}^\nclasses M_{ji}\right)$
		\ENDFOR
		\ENDWHILE
	\end{algorithmic}
\end{algorithm}

\section{Computation cost}
\label{sec:CompCost}
Recall that to tackle UHC, our approach comprises three steps (Sec.~3 in the main paper, second paragraph): ({\it i}) obtaining $\{p_i\}_i$ from $\mathbf{x}\in\mathcal{U}$ and $\{C_i\}_i$, ({\it ii}) estimating $q$ from $\{p_i\}_i$, and ({\it iii}) training $C_U$ from $\mathbf{x}$ and $q$.
The computation cost of different methods in the main paper differs only in step ({\it ii}), while it is the same for all methods in steps ({\it i}) and ({\it iii}). Focusing on ({\it ii}), standard distillation (Sec.~3.1) needs $\mathcal{O}(NL)$ to compute $q$ from $\{p_i\}_i$, while cross-entropy (Sec.~3.3) and matrix factorisation (Sec.~3.4) methods need to solve an optimisation problem, incurring much higher cost of $\mathcal{O}(tNL)$, where $t$ is the number of optimisation iterations. However, step ({\it ii}) is parallelisable for both cross-entropy and matrix factorisation methods, and it is a fixed cost irrelevant of classifier models. In contrast, the cost of training neural networks in step ({\it iii}) significantly overwhelms this fixed cost, thus in practice the difference is almost negligible.

\section{Complete results for sensitivity analysis}
\label{sec:SenAna}
\begin{figure}
	\centering
	\includegraphics[width=1\linewidth]{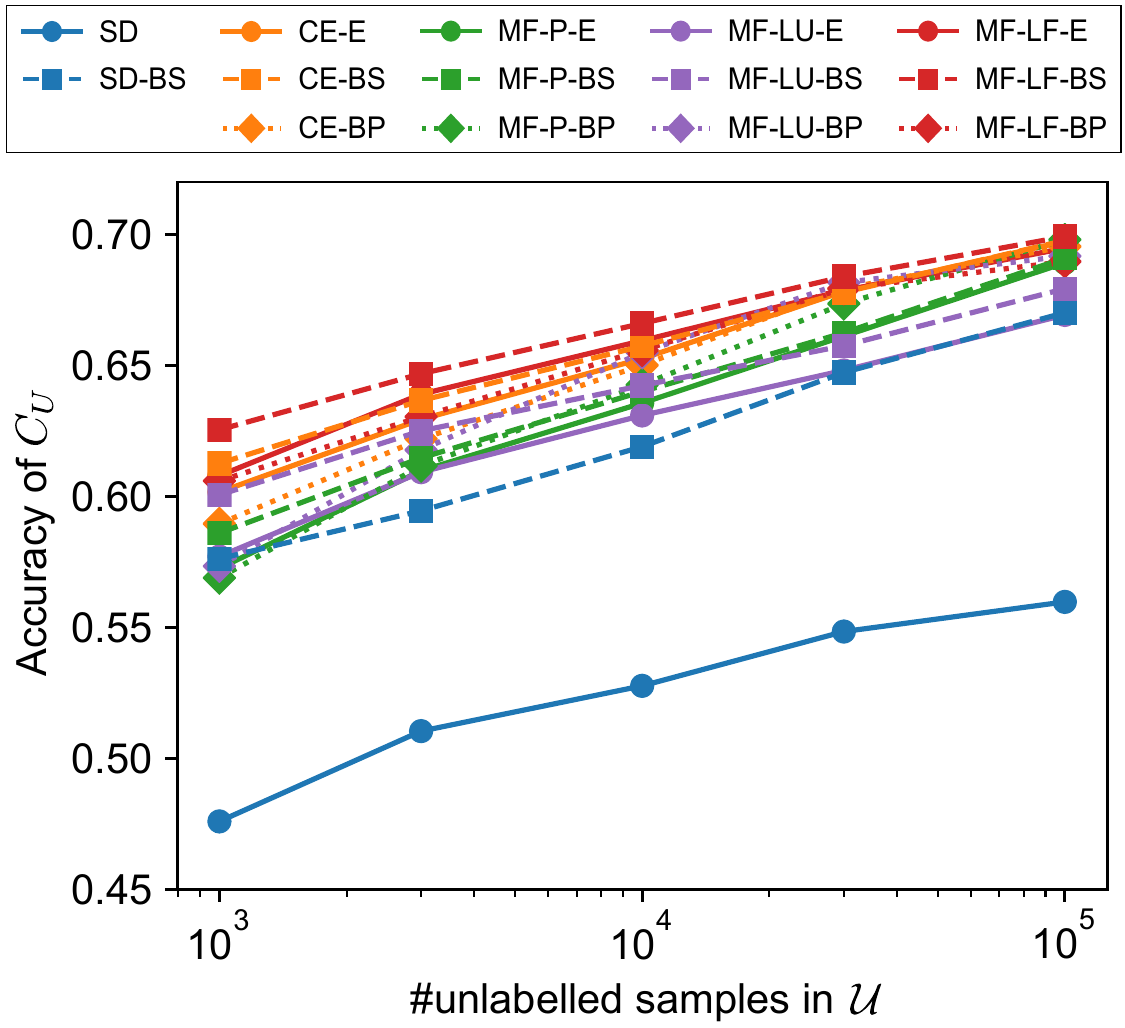}
	\caption{Sensitivity results on the size of the unlabelled set $\mathcal{U}$.}
	\label{fig:ablation_size_univ}	
\end{figure}

\begin{figure}
	\centering
	\includegraphics[width=1\linewidth]{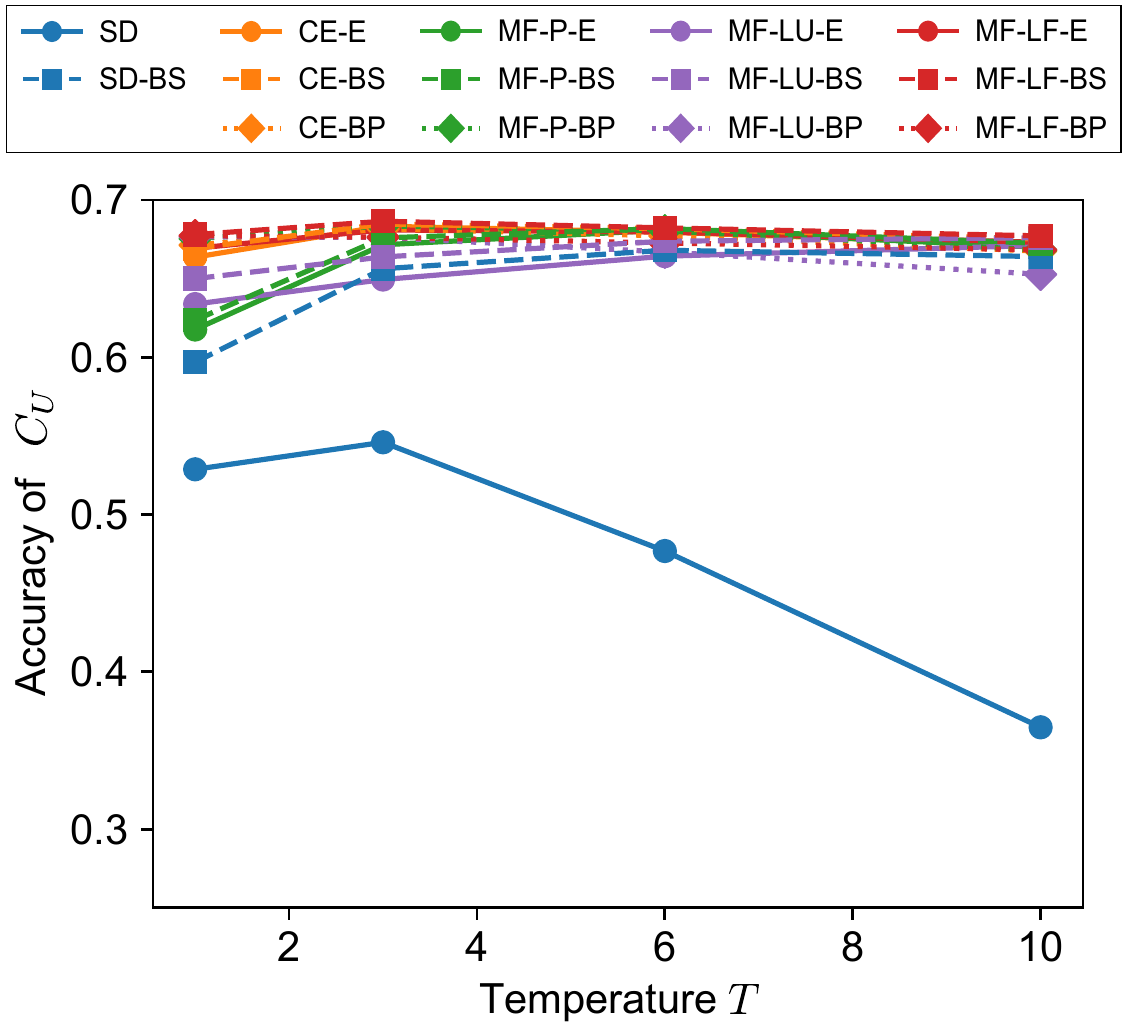}
	\caption{Sensitivity results on the temperature $T$.}
	\label{fig:ablation_temperature}	
\end{figure}

\begin{figure}
	\centering
	\includegraphics[width=1\linewidth]{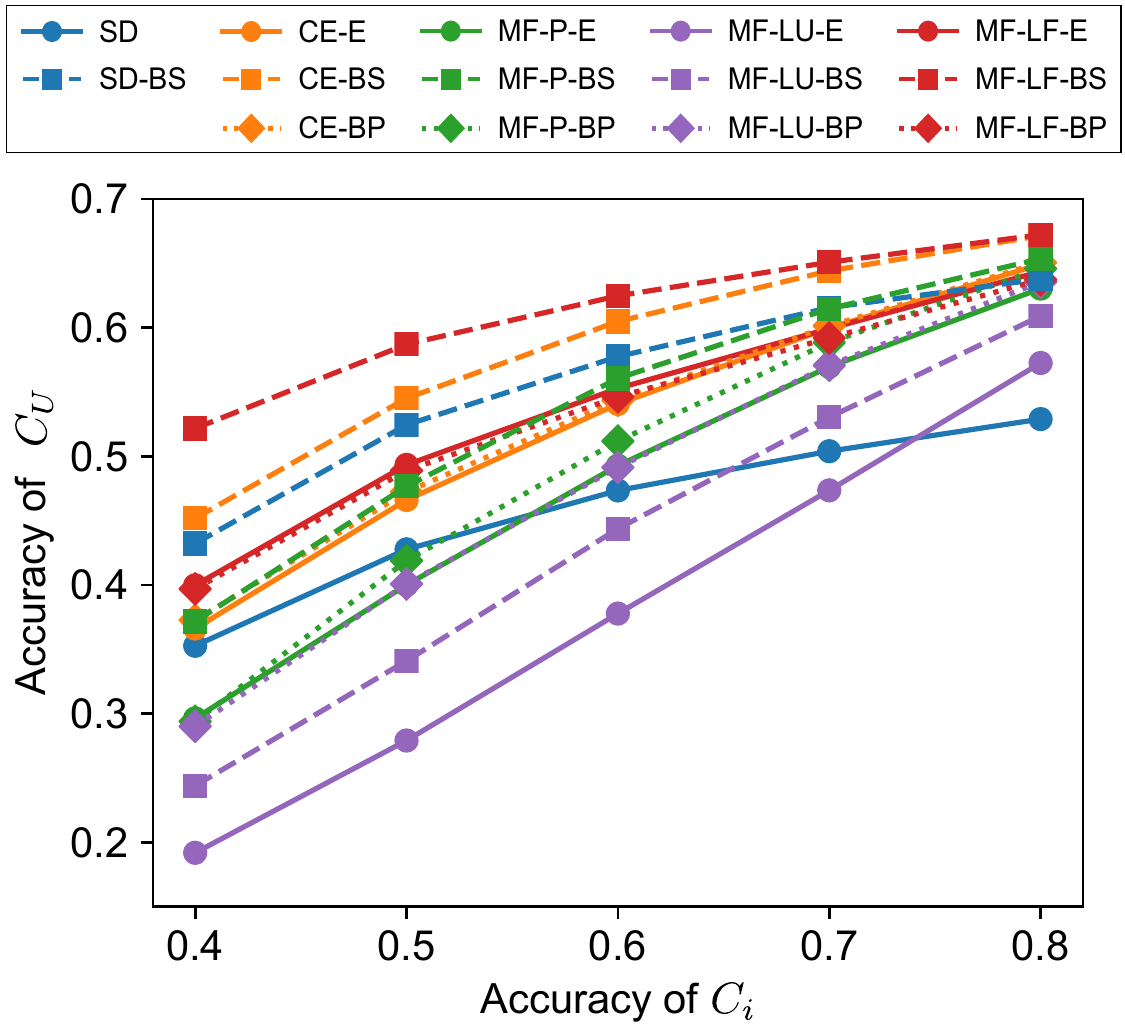}
	\caption{Sensitivity results on the accuracy of $C_i$.}
	\label{fig:ablation_accuracy_ci}	
\end{figure}

In this section, we provide the results of sensitivity analysis of all methods. Fig.~\ref{fig:ablation_size_univ} shows the sensitivity result for size of transfer set $\mathcal{U}$;  Fig.~\ref{fig:ablation_temperature} shows that of temperature $T$; and Fig.~\ref{fig:ablation_accuracy_ci} shows that of accuracy of $C_i$'s. Note that we use different legend style from the main paper to account for more methods.

\end{document}